\documentclass[sigconf,nonacm]{acmart}

\usepackage{multicol}

\usepackage[dvipsnames]{xcolor}

\usepackage{svg}
\usepackage{latexsym}
\usepackage{graphicx}
\usepackage{amsmath}
\usepackage{amsthm}
\usepackage{dsfont}
\usepackage{mathtools}
\usepackage{mathrsfs}
\usepackage[lofdepth,lotdepth]{subfig}
\usepackage{multirow}
\usepackage{booktabs}
\usepackage{lscape}
\usepackage{algorithm}
\usepackage{algorithmic}
\usepackage{array}
\usepackage{pifont}   
\usepackage{caption}
\usepackage{subcaption}

\usepackage{hyperref}
\usepackage{cleveref}

\usepackage{thmtools}
\usepackage{thm-restate}

\numberwithin{equation}{section}

\usepackage{soul}

\def\argmax{\mathop{\rm arg\, max}}

\newcommand{\bel}{\begin{eqnarray}\label}
\newcommand{\eel}{\end{eqnarray}}
\newcommand{\bes}{\begin{eqnarray*}}
\newcommand{\ees}{\end{eqnarray*}}
\newcommand{\bei}{\begin{itemize}}
\newcommand{\eei}{\end{itemize}}
\newcommand{\beiftnt}{\begin{itemize}\footnotesize}

\def\benu{\begin{enumerate}}
\def\eenu{\end{enumerate}}

\def\argmax{\mathop{\rm arg\, max}}

\def\complex{\mathop{{\rm I}\kern-.58em\hbox{\rm C}}\nolimits}

\def\mathbold{\boldsymbol} 

\def\btheta{\mathbold{\theta}}

\AtBeginDocument{%
  }

\setcopyright{acmlicensed}
\copyrightyear{2025}
\acmYear{2025}
\acmDOI{XXXXXXX.XXXXXXX}

\acmISBN{978-1-4503-XXXX-X/18/06}

\begin{document}

\title{Deep Reinforcement Learning for Ranking Utility Tuning \\
in the Ad Recommender System at Pinterest}


\author{Xiao Yang}
\authornote{Both authors contributed equally to this research.}
\authornotemark[0]
\affiliation{%
  \institution{Pinterest Inc.}
  \city{San Francisco}
  \state{California}
  \country{USA}
}
\email{xyang@pinterest.com}

\author{Mehdi Ben Ayed}
\authornotemark[1]
\affiliation{%
  \institution{Pinterest Inc.}
  \city{New York City}
  \state{New York}
  \country{USA}
}
\email{mbenayed@pinterest.com}

\author{Longyu Zhao}
\affiliation{%
  \institution{Pinterest Inc.}
  \city{San Francisco}
  \state{California}
  \country{USA}
}
\email{longyuzhao@pinterest.com}

\author{Fan Zhou}
\affiliation{%
  \institution{Pinterest Inc.}
  \city{San Francisco}
  \state{California}
  \country{USA}
}
\email{fanzhou@pinterest.com}

\author{Yuchen Shen}
\authornote{This work was done when Yuchen Shen was an intern at Pinterest.}
\affiliation{%
  \institution{Carnegie Mellon University}
  \city{Pittsburgh}
  \state{Pennsylvania}
  \country{USA}
}
\email{yuchens2@andrew.cmu.edu}

\author{Abe Engle}
\affiliation{%
  \institution{Pinterest Inc.}
  \city{San Francisco}
  \state{California}
  \country{USA}
}
\email{aengle@pinterest.com}

\author{Jinfeng Zhuang}
\affiliation{%
  \institution{Pinterest Inc.}
  \city{San Francisco}
  \state{California}
  \country{USA}
}
\email{jzhuang@pinterest.com}

\author{Ling Leng}
\affiliation{%
  \institution{Pinterest Inc.}
  \city{Seattle}
  \state{Washington}
  \country{USA}
}
\email{lleng@pinterest.com}

\author{Jiajing Xu}
\affiliation{%
  \institution{Pinterest Inc.}
  \city{San Francisco}
  \state{California}
  \country{USA}
}
\email{jiajing@pinterest.com}

\author{Charles Rosenberg}
\affiliation{%
  \institution{Pinterest Inc.}
  \city{San Francisco}
  \state{California}
  \country{USA}
}
\email{crosenberg@pinterest.com}

\author{Prathibha Deshikachar}
\affiliation{%
  \institution{Pinterest Inc.}
  \city{San Francisco}
  \state{California}
  \country{USA}
}
\email{pdeshikachar@pinterest.com}

\renewcommand{\shortauthors}{Yang et al.}

\begin{abstract}
The ranking utility function in an ad recommender system, which linearly combines predictions of various business goals, plays a central role in balancing values across the platform, advertisers, and users. Traditional manual tuning, while offering simplicity and interpretability, often yields suboptimal results due to its unprincipled tuning objectives, the vast amount of parameter combinations, and its lack of personalization and adaptability to seasonality.

In this work, we propose a general \textbf{D}eep \textbf{R}einforcement \textbf{L}earning framework for \textbf{P}ersonalized \textbf{U}tility \textbf{T}uning  (DRL-PUT) to address the challenges of multi-objective optimization within ad recommender systems. Our key contributions include: 1) Formulating the problem as a reinforcement learning task: given the state of an ad request, we predict the optimal hyperparameters to maximize a pre-defined reward. 2) Developing an approach to directly learn an optimal policy model using online serving logs, avoiding the need to estimate a value function, which is inherently challenging due to the high variance and unbalanced distribution of immediate rewards. 

We evaluated DRL-PUT through an online A/B experiment in Pinterest's ad recommender system. Compared to the baseline manual utility tuning approach, DRL-PUT improved the click-through rate by 9. 7\% and the long click-through rate by 7.7\% on the treated segment. We conducted a detailed ablation study on the impact of different reward definitions and analyzed the personalization aspect of the learned policy model.
\end{abstract}





\maketitle

\section{Introduction}

Recommender systems are crucial to the success of modern social networks, powering home feeds on platforms like Facebook, LinkedIn, TikTok, Instagram, Pinterest, to name a few~\citep{he2014practical,covington2016deep, jing2015visual,rethink22,liu2022monolith}. A personalized cascaded ranking layer often dictates the efficiency of content delivery, which directly impacts user satisfaction and the platform's business value. At a high level, such a ranking layer consists of two major components: 
\begin{itemize}
 \item \textbf{Prediction Models} that forecast a collection of organic engagement goals, e.g., content click-through rate (CTR)~\citep{cheng2016wide,guo2017deep,wang2017deep,zhou2018deep,zhai2024actions}, and monetization goals, e.g., ad conversion rate (CVR) ~\citep{lu2017apractical,shan2018combined};
 \item \textbf{The Utility Function} that integrates various predictions into a single utility score which serves as the primary metric to rank items. In practice, such a utility function is often a linear weighted sum of model predictions~\citep{cai2017real,bottou2013counterfactual}.
\end{itemize}

Modern machine learning models based on deep neural networks (DNN) achieve great success in predicting business goals, with CTR prediction being a prime example ~\citep{cheng2016wide,guo2017deep,wang2017deep,zhou2018deep,zhai2024actions}. Although accurately predicting these metrics is crucial, it is equally, if not more important, to balance each prediction in the utility score. In particular, during peak holiday periods such as Black Friday, the models must accommodate surges in engagement traffic and the influx of newly added items, which often present cold-start challenges. Tuning the utility score can provide immediate feedback that aligns with evolving business demands.

In this paper, we focus on the ad recommendation scenario, where the utility score must balance the value for the platform, advertisers, and users. Traditionally, the hyperparameters -- such as weights assigned to each business goal -- are determined by a cross-functional leadership team. These hyperparameters are typically static and require manual adjustments to align with evolving business priorities. This approach is classical and still very popular in industry today for its clarity and interpretability. However, it has become increasingly challenging in modern recommendation systems for the following reasons:

\textit{The tuning methodology is often unprincipled}. Modern recommendation systems typically optimize multiple objectives simultaneously. However, the mathematical formulation of these objectives during manual tuning is often unclear.

\textit{The total number of possible hyperparameter combinations is combinatorially explosive}. There are numerous items to recommend, each with its own objectives. Various engagement scores need to be predicted, each playing a significant role in user satisfaction. Furthermore, there are numerous business rules where each is associated with hyperparameters.

\textit{Lack of personalization and seasonal adjustment}. Manual tuning often results in a set of static hyperparameters: they remain unchanged in terms of both personalization, where the same hyperparameters are applied to all users, and seasonalization, where the same hyperparameters are not adjusted for different times or seasons.

Meanwhile, reinforcement learning (RL) algorithms, which can continuously update strategies based on user real-time feedback and aim to maximize expected rewards across various business scenarios, have become a natural fit for optimizing large-scale advertising systems in industrial settings.
As a consequence, RL has emerged as a transformative tool in digital advertising, addressing critical challenges in bid optimization~\citep{jin2018real,cai2017real,wu2018budget,rohde2018recogym,zhao2018deepbidding}, ads placement and display~\citep{zhao2018deep,zhao2020jointly,zhao2021dear,liao2022deep}, personalization and privacy~\citep{wang2018learning,xie2021hierarchical,timmaraju2023towards}, and budget / impression pacing~\citep{wei2023rltp}. We refer to~\citep{zhao2019deep} as a more comprehensive summary on this topic.

However, implementing RL in real-production utility tuning is never as simple as an off-the-shelf adoption. In this work, we propose a general \textbf{D}eep \textbf{R}einforcement \textbf{L}earning framework for \textbf{P}ersonalized \textbf{U}tility \textbf{T}uning (DRL-PUT), to address
the unique challenges of multi-objective optimization in an ad recommender
system at Pinterest. DRL-PUT employs an RL agent to predict the optimal hyperparameters. Upon receiving an ad ranking request from a user, the RL agent translates the user's features into a state representation. It then determines the optimal set of hyperparameters as its actions, guided by a learned policy aimed at maximizing customized rewards in various business scenarios.

DRL-PUT has significant advantages over traditional manual tuning. Customized rewards can mathematically articulate the tuning objectives, enabling the RL agent to adapt systematically to different desired business outcomes. The DNN-based policy model can effectively achieve personalization and adaptability to seasonality by exploiting the input personalization and contextual information, allowing significantly better decision-making quality.

DRL-PUT is quite different from the existing approaches \citep{feng2018learning,hu2018reinforcement,zhao2020whole,xie2021hierarchical} in industrial applications that treat the multi-scenario ranking task as a collaboration of different recommender systems through multi-agent reinforcement learning (MARL).

First, the RL agent in DRL-PUT functions independently of all other ranking models associated with the ranking utility function. This independence simplifies development and maintenance. In particular, the RL agent in DRL-PUT only adjusts the weights of the outputs of other ranking models, ensuring that the RL process does not require updates to these models. This is particularly advantageous for large-scale industrial online ad platforms like Pinterest, where modifying numerous models can be complicated.

Second, DRL-PUT is a policy based approach, making it computationally efficient in an online learning environment. As detailed in Section~\ref{section: action space}, we developed a method to discretize the action space into a manageable set and leveraged a DNN model-based policy gradient method. This approach allows us to learn an optimal policy directly, bypassing the need to estimate a value function, which is inherently challenging due to the high variance and unbalanced distribution of immediate rewards for each state in an online recommendation system~\citep{hu2018reinforcement}. Furthermore, at the serving time, the learned policy model only needs to be inferred once per request.

Our key value proposals include:

\begin{itemize}
\vspace{-0.1cm}
    \item We propose a general deep reinforcement learning framework to address the multi-objective ranking utility tuning problem. This approach provides an automatic, adaptive, and personalized solution. To the best of our knowledge, this is the first published work in this area.
    \item We validate our approach using Pinterest's ad ranking system in production and demonstrated that our model achieves significant performance gains in a real-world ad recommender system.
\vspace{-0.1cm}
\end{itemize}

Reinforcement learning, while theoretically robust and technically mature, faces challenges in web-scale deployments due to the added risk and complexity for product management.  To mitigate these challenges, we are actively developing observability, logging, and alerting systems to facilitate the deployment and understanding of the online behaviors of DRL-PUT.

\section{Related Works}
\label{section: related works}

In this section, we briefly review several categories of works that are closely related to ours.


\subsection{RL-based Online Advertising} 

In recent years, RL based methods have gained great success in online advertising on various industrial platforms. In~\citep{cai2017real}, the authors modeled the state transition via auction competition and derived the optimal policy to bid for each impression with a model-based Markov Decision Process (MDP), while in~\citep{wu2018budget} the authors transformed the original bidding process into $\lambda$ -regulating and proposed a model-free MDP to derive the optimal policy for bidding. In order to model the interactions of all merchants in the bidding process, in~\citep{jin2018real} the authors proposed a Distributed Coordinated Multi-Agent Bidding solution using MARL. These methods focus on applying RL to bidding optimization, while ours focus on ad recommendation.

Another line of research is value-based RL approaches in online advertising. For instance, in~\citep{zhao2021dear}, the authors proposed a Deep Q-network (DQN) architecture for online advertising to simultaneously determine whether to display an ad and which and where to display it. In~\citep{wei2023rltp}, the authors proposed RLTP to learn a value function $Q(s, a)$ based on a tailored reward function, in order to jointly optimize impression count and performance metrics such as CTR. In~\citep{wang2018learning}, the authors formulate the problem as a constrained MDP with per-state constraints and proposed a two-level RL strategy. The key differences between~\citep{wei2023rltp,zhao2021dear, wang2018learning} and our work is that \citep{wei2023rltp,zhao2021dear, wang2018learning} are value-based methods that focus on finding the optimal value function, from which the optimal policy can be derived, while ours is a policy-based method that learns the optimal policy directly. This policy-based method significantly improves efficiency in handling large action spaces in a large-scale online advertising platform.



\subsection{RL-based recommender system and ranking}

Another closely related area to our work involves RL based ranking and recommender systems.

In ~\citep{ayed2025recomindreinforcementlearningframework}, RecoMind tackles RL-based ranking and recommendation at web scale, introducing a simulator-based framework that efficiently optimizes session-based objectives across massive action spaces. In~\citep{feng2018learning}, to achieve optimal balance among recommender systems targeting different business metrics, the authors propose a multi-agent Recurrent Deterministic Policy Gradient method (MA-RDPG) with continuous action space. In their model, private actors operate within each optimization domain and work collaboratively through a shared action-value function (critic). In contrast, our work employs a single RL agent with discrete action space to directly learn the weights in a ranking utility function to balance different factors.

In ~\citep{hu2018reinforcement,zhao2020whole}, the authors studied the problem of multi-step ranking during a shopping/search session. In \citep{hu2018reinforcement} they formulated the problem as a model-based MDP within each search session and proposed a deterministic policy gradient method, while in \citep{zhao2020whole} they considered the joint learning of multiple sequential recommenders within a shopping session. Specifically, they proposed a MARL framework with an actor-critic method where each recommendation agent (RA) at different timestamp in one session is activated sequentially while all actors share a global critic. Our method simplifies the problem formulation, by concentrating solely on a single step through the ranking utility function. Our online experiments have demonstrated that this simplification is sufficiently effective.

Another loosely related work is \citep{xie2021hierarchical}, where the authors proposed a hierarchical RL framework (HRL-Rec): the low-level RL agent recommends a channel list, and the high-level RL agent recommends an item list within each channel. As a comparison, our approach targets a different area, achieving improved ranking and recommendations through personalized weights in a ranking utility function.

\section{Problem Formulation}\label{section: method}

\subsection{Preliminaries and Notations}
We use lowercase letters $a,s,r$ to denote action, state, and reward, respectively. 
We use curly letters $\mathcal{A}, \mathcal{S}, \mathcal{F}$ to denote the action space, state space, and feature space, respectively. Let $\pi$ denote the policy and $\pi(a|s,\btheta)$ be the probability of taking action $a$ in state $s$ where $\pi$ is parameterized by $\btheta$.

\subsection{Ranking Utility Formula}
\label{section: objective}

In an ad ranking system, the ranking utility score is used to determine the best ad to display for a given ad request, which has to be carefully tuned to maximize a set of key business metrics, balancing the interests of the platform, the advertisers, and the users to achieve desirable trade-offs.

A common practice is to represent the ranking utility $\mathbf{U}$ for a given ad item in a weighted sum form:
\begin{equation}
\label{formula: utility}
\begin{aligned}
    \mathbf{U} :=& \ \mathbf{1} \{ Estimated\_Revenue \geq b \} \cdot ( Estimated\_Revenue \\
    &+ Estimated\_User\_Value) \\
    =& \ \mathbf{1} \{ Estimated\_Revenue \geq b \} \cdot \big( \ p(optimized\_action) \cdot bid \\
    &+ \sum_{i=1}^{n-1} p(engagement\_action_i) \cdot w_i \big)
\end{aligned}
\end{equation}
where:
\begin{itemize}
    \item $p(optimized\_action)$: the probability of the action that an ad aims to optimize. An optimized action can be click, conversion, impression, etc. depending on the ad campaign type.
    \item $bid$: advertiser's bid for the optimized action.
    \item $p(engagement\_action_i)$: the probability of an engagement action. An engagement action can be click, click longer than $t$ seconds, conversion, etc.
    \item $w_i$: weight of the corresponding engagement action.
    \item $b$: threshold on $Estimated\_Revenue$, also referred to as reserve price.
    \item $n$: total number of hyperparameters including all $w_i$ and $b$.
    \item $\mathbf{1} \{\cdot\}$: indicator function that returns 1 if the condition in the curly brackets is satisfied; it returns 0 otherwise.
\end{itemize}

The term $Estimated\_Revenue$ refers to the projected monetary value that can be generated by displaying an ad. A formal definition for our use case can be found in~\eqref{reward: revenue}.
The term $Estimated\_User\_Value$ represents the estimated value provided to users through various measurable engagement actions. For example, $p(click)$ represents the likelihood that a user shows interest in an ad and clicks on it; $p(click30)$ represents the likelihood that a user spends more than $30$ seconds on the landing page of an ad after clicking on it; $p(conversion)$ represents the likelihood of whether a user purchases an item. Intuitively and empirically, assigning a higher weight to $p(engagement\_action_i)$ in the ranking utility is likely to increase the overall frequency of $engagement\_action_i$.

As we can see, in this context, optimizing business top-line metrics $\mathcal{M}$ translates to selecting the optimal hyperparameters $\mathcal{A}:=\big\{b, \{w_i\}_{i=1}^{n-1}\big\}$ for a given request, utilizing all available information $\mathcal{F}$ upon receiving an ad request. 

By formulating this utility tuning problem in RL terminology, 
\begin{itemize}
    \item the \textbf{action space} is the combinations of variables in $\mathcal{A}$;
    \item the \textbf{state space} is the representations based on $\mathcal{F}$;
    \item the \textbf{reward} is a function of $\mathcal{M}$.
\end{itemize}
In this paper, we only consider a one-step action instead of a trajectory of actions for simplicity.

\subsection{Action Space}\label{section: action space}

As we mentioned in Section~\ref{section: objective}, the actions correspond to choosing $\mathcal{A}$ in the ranking utility formula \eqref{formula: utility}. By default, we have $b \in \mathbb{R_+}$ and $w_i \in \mathbb{R_+}$ for $i=1,\dots,n-1$, where $\mathbb{R_+}$ denotes the set of positive numbers. In our case $n$ easily goes to double digits. 

Instead of utilizing the continuous action space $\mathbb{R}_+^n$, we opted for a discretized one. This is because we cannot obtain a sufficiently large training dataset for a continuous action space without a simulation system; see Section~\ref{section: data} on data collection. We observe that our model failed to converge during the training stage when the number of training examples is much smaller than the continuous action space.

We then explored reducing the action space by discretizing $\mathcal{A}$ in the following way: for each $w_i$ (the same for $b$, omitted for simplicity), we set a predefined range as $[w_{i,\min}, w_{i,\max}]$ based on some prior information, and then partition this range into a set of $m$ equally spaced values, $w_i \in \{w^{(1)}_i, w^{(2)}_i, \dots, w_i^{(m)}\}$ with $w_{i}^{(1)} = w_{i,\min}$ and $w_{i,\max} = w_i^{(m)}$. In our case $m = 10$ which is a tuned hyperparameter. This discretization successfully reduced the total action space from $\mathbb{R}_+^n$ to a manageable discretized one of size $m^n$.

We further reduce the action space cardinality by "correlating" $w_i$'s: we group $w_i$'s into subsets if the corresponding $p(engagement\_action_i)$ are semantically related. For example, $p(click)$ and $p(click30)$ are physically related, but are not related to $p(conversion)$. Once $w_i$ and $w_j$ are grouped together, their values shall always be the same $i^{th}$. In this way, we further reduce the action space from a set of $m^n$ combinations to $g^n$, where $g$ is the number of groups and $g = 3$ in our case.

Intuitively, reducing the action space makes the learning process much simpler. Moreover, such a discretization method enables us to adopt RL approaches to directly predict actions given states without consulting a state-action value function. This is particularly crucial in the online learning production environment, where evaluating the value of all possible states/actions is computationally expensive.

 \subsection{State Representation}
 
We use $\mathcal{F}$ to denote the available features when receiving an ad request before taking an action, which can be divided into the following categories:
\begin{enumerate}
    \item \textbf{User profile}: information about the user profile, such as \textit{age, gender, metro area}, etc;
    \item \textbf{User activities}: user historical actions and counts either on-site (e.g., \textit{clicked items}) or off-site (e.g., \textit{purchased items});
    \item \textbf{Contextual information}: information about the ad request, such as \textit{hour of the day, day of the week, IP country}, etc.
\end{enumerate}

Note that for an advertising system, the demand (i.e. the overall spending budget of all ads in inventory) and supply (i.e. activities of all users that can attribute to an ad) information at a given time also play an important role in determining the optimal choice of $\mathcal{A}$ for a given ad request. However, such information is either hard to measure or infeasible to be obtained during serving time. Therefore, we do not use them in this work. 

\subsection{Reward Function}\label{section: reward definition}

The reward function guiding the learning process is a crucial component of our approach. We use $r$ to denote the generic reward, encoding the business metrics that we aim to optimize, such as revenue, CTR, etc. Similar to the definition of the utility function~\eqref{formula: utility}, $r$ consists of two major components: 
\begin{equation}
\label{reward: generic}
    r := Estimated\_Revenue + Estimated\_User\_Value.
\end{equation}

\subsubsection{$Estimated\_Revenue$}
Due to the fact that ad campaigns aim to optimize different goals, for example, click-through campaigns aim to maximize click volume, whereas conversion campaigns focus on boosting the number of conversions, we define the \textbf{revenue component} aligning to these diverse business objectives:
\begin{equation}
\label{reward: revenue}
\begin{aligned}
   & Estimated\_Revenue \\
   & := 
\begin{cases} 
p(click) \cdot  bid_{ctr} & \text{for}~clickthrough~campaign; \\
p(conversion) \cdot bid_{conv} & \text{for}~conversion~campaign; \\
bid_{imp} & \text{for}~impression~campaign;
\end{cases}
\end{aligned}
\end{equation}
with $bid_{ctr}$, $bid_{conv}$, and $bid_{imp}$ being the advertiser's bid price for each campaign type.

\subsubsection{$Estimated\_User\_Value$}
The \textbf{user value component} of reward is the weighted sum of the estimated probabilities:

\begin{equation}
\label{reward:user}
\begin{aligned}
 Est&imated\_User\_Value \\
& := \alpha \cdot p(click) + \beta \cdot p(click30) + \gamma \cdot p(conversion).
\end{aligned}
\end{equation}
Here $\alpha$, $\beta$, and $\gamma$ are a small number of meta hyperparameters in the reward definition with different values depending on the campaign type. 

An interesting phenomenon we observed is that applying min-max batch normalization to each estimated engagement probabilities in~\eqref{reward:user} can significantly improve the stability and convergence speed of training. For $N$ samples within a batch $\mathcal{B}$, $p(engagement\_action_i)$ is normalized by first subtracting the minimum value in the batch and then divided by the range (maximum - minimum) so that the normalized value fits in $[0, 1]$.

\section{DRL-PUT: Architecture and Training}\label{section: policy gradient}
In this section, we present the architecture of the policy model and the policy gradient method for training in DRL-PUT. 

\subsection{High-level Design Choice} 
Unlike traditional action-value methods (see Chapters 2 and 3 in ~\citep{sutton2018reinforcement}), where selected actions are based on estimated action values, policy gradient methods learn a parametrized policy that directly selects actions without consulting a value function. This simplicity is crucial for our purpose because it is prohibitively difficult to estimate a value function in an ad recommender system, due to the high variance and unbalanced distribution of immediate reward for each state~\citep{hu2018reinforcement}. Moreover, it is infra-costly in practice due to the large action space.

There are two types of generic policy gradient methods that have been extensively studied and widely adopted: REINFORCE~\citep{williams1987reinforcement,williams1992simple} and Actor-Critic methods~\citep{witten1977adaptive,barto1983neuronlike,sutton1984temporal,Degris2012}. The REINFORCE method improves its policy by learning from the rewards it receives. It adjusts the policy to increase the likelihood of actions that lead to higher rewards. On the other hand, Actor-Critic methods learn approximations to both policy and value functions. The term "actor" refers to the learned policy and the term "critic" refers to the learned value function. In this work, we experimented with both methods and adopted REINFORCE. The comparison details are in Section~\ref{section: offline exp}.

\subsection{Architecture of the Policy Model}\label{section: model arch}

Our policy model $\pi:\mathcal{S} \rightarrow \mathcal{P}(\mathcal{A})$ maps the state space $\mathcal{S}$ to the probability distribution $\mathcal{P}(\mathcal{A})$ over the action space $\mathcal{A}$.


As we mentioned in Section~\ref{section: action space}, \textbf{action space discretization} allows us to formulate the policy into $\argmax_{a\in\mathcal{A}} \pi(a|s, \btheta)$ for each pair $(s,a)$ through a deep neural network with weights $\btheta\in\mathbb{R}^d$. 

\begin{figure}
    \centering
    \includegraphics[width=\linewidth]{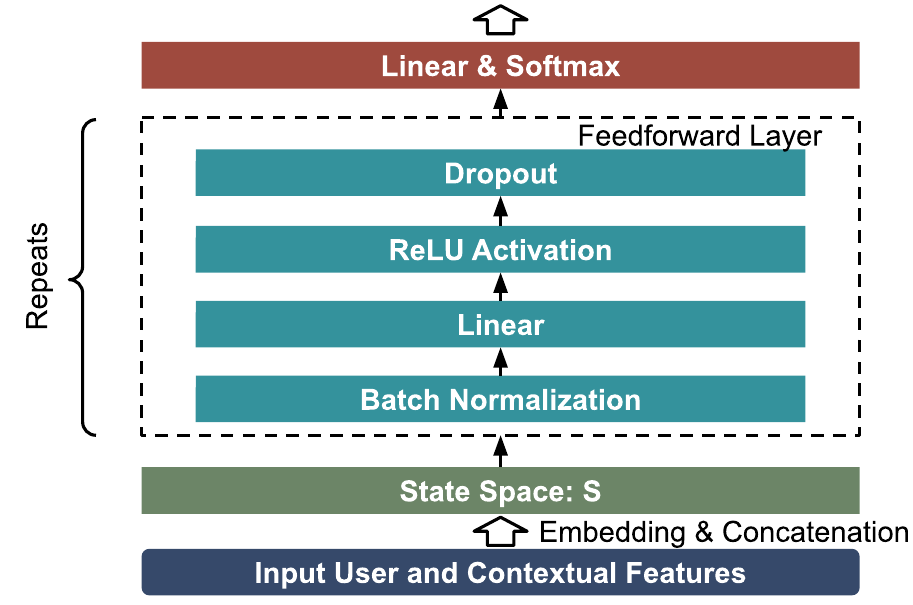}
    \vspace{-0.3cm}
    \caption{The architecture of the policy model.}
    \label{fig:model diagram}
    \vspace{-0.3cm}
\end{figure}

We build our policy model based on multilayer perceptrons (MLP)~\cite{rosenblatt1958perceptron}, which learns to select the best action $a_i$ given the current state $s_i$. Specifically, we first feed each categorical feature into a corresponding embedding layer to obtain representation vectors, then concatenate them with numerical features and dense vector features. Then this representation is fed into a sequence of batch normalization layer~\citep{ioffe2015batch}, linear layer, and non-linear layer with ReLU~\citep{nair2010rectified} activation. The final output comes from a softmax layer which represents the probability distribution over the discretized action space $\mathcal{A}$. The action with the highest probability will be selected for each state input. The architecture of the policy model is illustrated in Fig.~\ref{fig:model diagram}.

\subsection{Policy Gradient Algorithm}
With the policy model parameterized by weights $\btheta \in \mathbb{R}^d$, our policy gradient method is based on the REINFORCE algorithm \citep{williams1987reinforcement,williams1992simple}.

In the REINFORCE algorithm, a discount factor $\gamma \in [0, 1]$ is usually used, which aims to balance the trade-off between the immediate reward and the future reward. We set $\gamma=0$ purposely to imitate a completely myopic agent that only considers the immediate reward observed after completing a trajectory. Especially, we calculate the policy gradient update through a slightly modified version of the original stochastic gradient ascent in the original REINFORCE algorithm, where each episode consists of just a single step and a batch of $B$ ad requests over a short period of time. Let $\eta >0$ be the step size, for each episode $t$ we set
\begin{equation}
\label{formula: gradient ascent}
\btheta_{t+1} = \btheta_t + \eta \cdot \frac{1}{B} \cdot \sum_{i=1}^B r^{(i)}_t \cdot \nabla_{\btheta} \log \pi(a^{(i)}_t|s^{(i)}_t,\btheta_t).
\end{equation}
Note that in the second term on the right-hand side of \eqref{formula: gradient ascent},  we update the policy gradient using the average of a batch consisting of $B$ users, instead of relying on a single user's state representation and action preference. This mini-batch averaging strategy reduces the variance of the gradients and consequently leads to faster convergence and more stable parameter updates.

To explain the algorithm in detail: for each episode $t$, when the user's ad request arrives, we transform all user and contextual features into the state representation $s_t$, then we feed it into the policy model $\pi$ to obtain the predicted probability distribution over all actions for the given request. Once a batch is collected, we use the gradient ascent formula \eqref{formula: gradient ascent} to update the weights of the policy model $\btheta$ to increase the logarithmic likelihood of the logged action weighted by the corresponding immediate reward $r_t$. The algorithm is summarized as in Algorithm~\ref{alg:reinforce}.
 \begin{figure}
    \centering
    \includegraphics[width=1.0\linewidth]{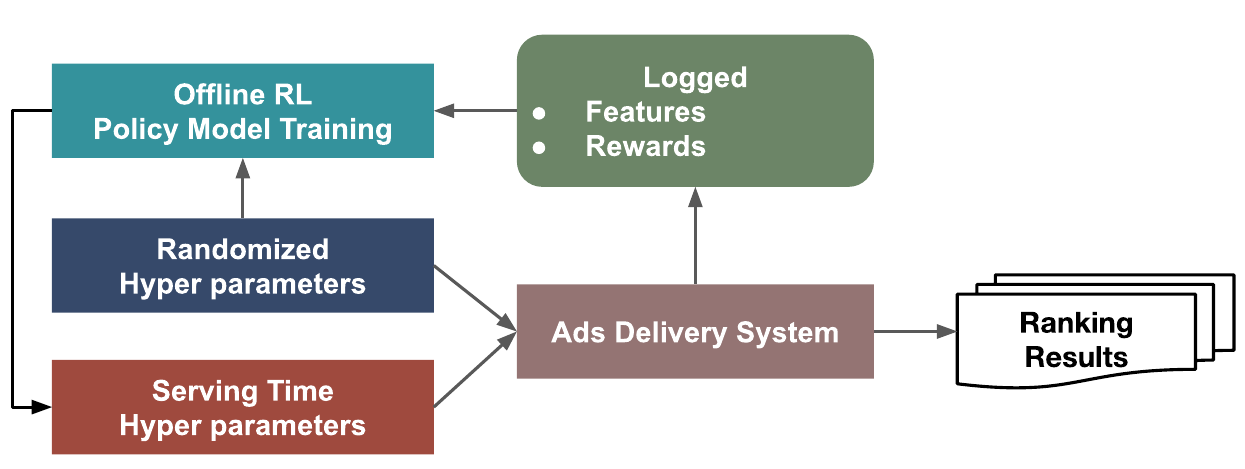}
    \vspace{-0.3cm}
    \caption{Workflow of data collection and model serving.}
    \vspace{-2mm}
    \label{fig:diagram}
\end{figure}

\begin{algorithm}
\caption{Policy Gradient Training of DRL-PUT }
\label{alg:reinforce}
\begin{algorithmic}
\STATE \textbf{Input:} Policy model $\pi(a|s, \btheta)$, reward function $r$
\STATE \textbf{Initialize:} Step size $\eta> 0$, parameters $\btheta\in\mathbb{R}^d$
\FOR{episode: $t=1,~2,~3,~\dots$}
    \STATE With each batch of state representations $\{s^{(i)}_t\}_{i=1}^{B}$:
    \STATE Compute probability of logged action $\{a^{(i)}_t\}_{i=1}^{B}$ via $\pi(a|s,\btheta)$;
    \STATE Compute immediate rewards $\{r^{(i)}_t\}_{i=1}^{B}$;
    \STATE Compute the averaged log loss for the batch
    \STATE  $$L(\btheta) = -\frac{1}{B} \sum_{i=1}^{B} r^{(i)}_t \cdot \log \pi(a^{(i)}_t | s^{(i)}_t, \btheta);$$
    \STATE Compute the policy gradient w. r. t. $\btheta$: $\nabla_{\btheta} L(\btheta)$;
    \STATE Update model parameters: $\btheta \leftarrow \btheta - \alpha \cdot \nabla_{\btheta} L(\btheta)$;
\ENDFOR
\end{algorithmic}
\end{algorithm}

\section{Data Collection and Training Details}\label{section: data}

In this section, we outline the data collection and continuous model training procedures, as illustrated in Figure~\ref{fig:diagram}. To initiate model learning, we first gather training data, which comprises triplets of state, action, and reward, denoted $\langle s,a,r \rangle$. This is accomplished by reserving a small percentage $(x\%)$ of online traffic during serving time, where we sample the action $a$ according to a behavioral policy $\pi^B$. To minimize potential risks in the production system, we restrict $x\%$ to be no more than $0.5\%$ at the request level.

The exact form of the behavior policy $\pi^B$ is heavily related to the action space $\mathcal{A}$. As described in Section~\ref{section: action space}, we experimented with various options and eventually adopted a discretized grouped action space with $g^n = 10^3$ possible actions in total. The behavior policy $\pi^B$ we choose is a uniform distribution over the discretized action space $\mathcal{A}$: $\pi^B \sim \mathbf{U}(\mathcal{A})$, from which we sample the action at serving time for each request state. 

Another choice of $\pi^B$ we tried is a discrete analogue of Gaussian
distribution: $\pi^B \sim \lfloor \mathcal{N}(\mu, \sigma^2) \rfloor$, where $\mathcal{N}(\mu, \sigma^2)$ denotes Gaussian distribution and $\mu$ is selected as the hyperparameters in the ranking utility formula \eqref{formula: utility} already adopted in production, and $\sigma$ is selected such that all possible actions fall within $2\sigma$ from $\mu$. Here, $\lfloor \cdot \rfloor$ denotes the operation to select the closest discretized action.

However, we empirically observed that using such a behavior policy to train models results in insufficient exploration: There are too many sampled actions for training concentrated around the production hyperparameters $\mu$, and much fewer examples that are far away. Consequently, the model lacks sufficient exploration to enhance its decision-making strategy across the entire action space.


\section{Experiments}\label{section: experiments}
We conduct both offline and online experiments on real-world datasets from Pinterest, which has hundreds of millions of monthly active users. 

We aim to address the following questions: 
\begin{enumerate}
    \item Does DRL-PUT yield positive results? Specifically, does the model training process converge? Can the model effectively improve on-line business metrics?
    \item Does DRL-PUT effectively leverage user and contextual features to achieve personalization?
\end{enumerate}


\subsection{Offline Experiments}\label{section: offline exp}
RL for utility tuning introduces significant design ambiguities. The broad ranges of choices, such as defining action space, state representation, reward function, behavior sampling, and training algorithms. Our experiments with various configurations revealed that the model can fail to learn under certain conditions. Furthermore, the lack of clear offline evaluation methods complicates the iterative development process. To address this, we propose two offline evaluation metrics: $Diversity$ and $Relative\_Gain$.

\textit{Diversity} measures how dominant of the most frequent predicted actions $a^*$ for various states $s$. Formally, it is defined as: 
\begin{equation}
\begin{aligned}
     Diversity := 1 - N_{a^*}/N_{total}
\end{aligned}
\end{equation}
where $N_{a^*}$ denotes the frequency of the most frequently predicted action $a^*\in \mathcal{A}$ by the model and $N_{total}$ is the total number of training samples. We introduce a simple threshold $t$, where a model is deemed failure if $Diversity < t$.

\begin{table*}[h]
    \centering  
    \caption{Model Offline Behaviors Under Various Configurations. A checkmark (\ding{51}) indicates success.}
    \vspace{-3mm}
    \label{tab:offline}
    \begin{tabular}{ccccccccccc}
        \toprule 
        \textbf{Configuration} & \multicolumn{3}{c}{\textbf{Action Space}} & \multicolumn{2}{c}{\textbf{Behavior Logging}} & \multicolumn{2}{c}{\textbf{Training Algorithm}}  & \multicolumn{2}{c}{\textbf{Evaluation Metric}} \\
        & $\mathbb{R}_+^n$ & $m^n$ & $g^n$ & Gaussian & Uniform & Actor-Critic & REINFORCE & $Diversity$ & $Relative\_Gain$ \\
        \midrule
        1 & \ding{72} &  &  & \ding{72} &  & \ding{72} &  & \ding{55} & \ding{55}  \\
        2 & \ding{72} &  &  &  & \ding{72} & \ding{72} &  & \ding{55} & \ding{55}  \\
        3 & & \ding{72} &  & \ding{72} &  & \ding{72} &  & \ding{55} & \ding{55}  \\ 
        4 & & \ding{72} &  &  & \ding{72} & \ding{72} &  & \ding{55} & \ding{55}  \\
        5 & &  & \ding{72} &  & \ding{72} & \ding{72} &  & \ding{55} & \ding{55}  \\ 
        6 & &  & \ding{72} &  & \ding{72} &  & \ding{72} & \ding{51} & \ding{51}  \\ 
        \bottomrule
    \end{tabular}
\end{table*} 

\textit{Reletive\_Gain} quantifies the superiority of a candidate policy over behavior policy, weighted by reward $r$. It is defined as:
\begin{equation}
Relative\_Gain := \sum_{i=1}^{N_{total}} r_i \cdot \big( \pi (a_i | s_i, \theta) - \pi^B (a_i | s_i) \big)
\end{equation}
Clearly, a model is deemed a failure if $Relative\_Gain < 0$.

We explored various configurations for action space, behavior logging, training algorithms, and summarized model's behavior in Table~\ref{tab:offline}. As shown in the table, the careful selection of design choices is crucial for successful learning of policy models.

\subsection{Online A/B Experiments}
\subsubsection{Metrics}
Online experiments evaluate key business metrics, including revenue, impression (measuring the total ad impression), CTR, CTR30 (the proportion of clicks lasting longer than 30 seconds relative to total ad impressions), and CVR.

\subsubsection{Production Traffic}
\label{section: prod online exp}
We present the results of online A/B experiments with high production traffic volume in Table~\ref{tab:lcb}. We observed significant improvements in metrics both platform-wide and within the treated segment.

\begin{table}[h]  
    \centering  
    \caption{Online A/B Experiments Results. (* means statistically insignificant)}
    \vspace{-3mm}
    \label{tab:lcb}
        \resizebox{\columnwidth}{!}{
        \begin{tabular}{l c c c c c}
            \toprule
            & \textbf{Revenue} & \textbf{Impression} & \textbf{CTR} & \textbf{CTR30} & \textbf{CVR} \\
            \midrule
            Platform & +0.27\% & \textbf{+0.02\%}* & +1.62\% & +1.03\% & +0.67\% \\
            Treated Segment &\textbf{-0.16\%}* &\textbf{-0.08\%}* & +9.71\% & +7.73\% & +1.26\% \\
            \bottomrule
        \end{tabular}
        }
  \vspace{1mm} 
  \begin{flushleft}
  \small Note: 0.5\% increase in CTR is considered as a substantial gain.
  \end{flushleft}
\end{table}     

\subsubsection{Ablation Study for the Reward Function}
We conducted several additional online A/B experiments to evaluate the impact of varying the definition of rewards on business metrics.

\textbf{R0}: This is the reward function explained in Section~\ref{section: reward definition} and used in online A/B experiments described in Section ~\ref{section: prod online exp}. 
The coefficient values $\langle \alpha, \beta, \gamma \rangle$ are selected as $\langle 1.0, 0.5, 0.0 \rangle$, $\langle 0.1, 0.4, 0.5 \rangle$, and $\langle 0.0, 0.0, 0.0 \rangle$ for click-through, conversion, and impression campaigns, respectively.

\textbf{R1}: The reward is defined as $r := Estimated\_Revenue$, omitting the $Estimated\_User\_Value$ part.

\textbf{R2}: The reward is defined as $r := Estimated\_Revenue + \alpha \cdot p(click) + \beta \cdot p(click30)$, where all campaign types share the same value of $\alpha$ and $\beta$.

Table~\ref{tab:ablation1} compares \textbf{R0}, \textbf{R1}, and \textbf{R2}. \textbf{R1} has a clear positive impact on revenue but generally at the expense of user engagement. Specifically, CTR decreased significantly from $+9.71\%$ to $-0.74\%$, and CTR30 also dropped notably from $+7.73\%$ to $-3.81\%$. 
These changes in online metrics highlight the inherent trade-off between revenue and user satisfaction in
a pure revenue-focused strategy. Conversely, \textbf{R2} enhances both
CTR and CTR30 metrics, but at the cost of decreased revenue. 
In this scenario, more engaging ads that generate less immediate revenue tend to prevail, negatively impacting overall revenue.
Interestingly, CVR showed a slight improvement compared with \textbf{R1}, probably due to incorporating $p(click)$ and $p(click30)$ in the reward function. While still negative, this suggests potential for optimizing the click-conversion balance in the reward function.

\begin{table}[h]  
    \centering  
    \caption{Comparing R0, R1 and R2 across all campaigns (treated segment).}
    \vspace{-3mm}
    \label{tab:ablation1} 
        \begin{tabular}{l c c c c c}
            \toprule
            & \textbf{Revenue}& \textbf{Impression} & \textbf{CTR} & \textbf{CTR30} & \textbf{CVR} \\
            \midrule
            \textbf{R0} &  -0.16\% & -0.08\% & +9.71\% & +7.73\% & +1.26\% \\
            \textbf{R1} & +0.23\% & -0.64\% & -0.74\% & -3.81\% & -6.29\% \\
            \textbf{R2} & -1.29\% & -1.04\% & +11.0\% & +11.4\% & -1.05\% \\
            \bottomrule
        \end{tabular}
\end{table}  

\begin{figure*}
    \centering
    \includegraphics[width=\linewidth]{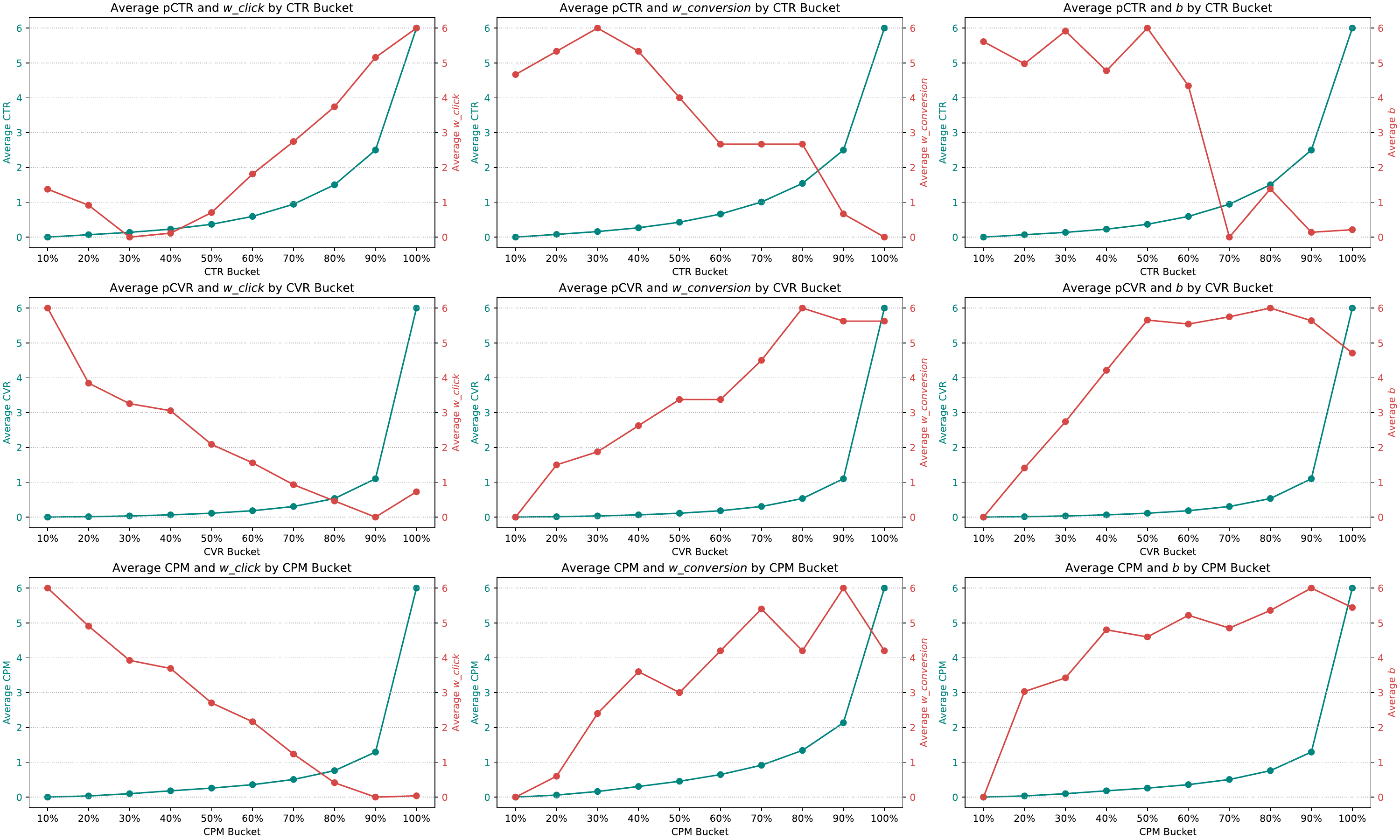}
    \vspace{-5mm}
    \caption{Average of the predicted hyper parameters within each bucket. Top: learned weights by CTR bucket. Middle: learned weights by CVR bucket. Bottom: learned weights by CPM buckets. Note we hide the absolute value of CTR / CVR / CPM and the learned weights, but the scale keeps linear on the y axis in each figure.}
    \label{fig:visualization}
    \vspace{-5mm}
\end{figure*}

\textbf{R3}: This configuration customizes the reward functions for different campaign types, motivated by \textbf{R2}'s performance on impression campaigns. As shown in Table ~\ref{tab:ablation2}, while \textbf{R2} increased CTR by 15.87\% for these campaigns, it also decreased revenue and impressions by 1.43\% and 2.41\%, respectively.
The significant drop in impressions does not align with the objectives of impression campaigns. Therefore, for these campaigns, in particular, we have used $Estimated\_User\_Value$ and instead scaled the revenue reward by a magnitude equivalent to $Estimated\_User\_Value$ used in other campaigns.  This adjustment compensates for the potential reduction in the magnitude of the total reward. As shown in Table~\ref{tab:ablation2}, this approach effectively mitigates the previously observed revenue and impressions losses in impression campaigns, while minimizing reductions in CTR and CTR30. Overall, this variant demonstrates that adjusting the reward
composition for different campaign types can effectively address the inherent
trade-offs. 

\begin{table}[h]  
    \centering  
    \caption{Comparing R2 and R3 on impression campaigns (treated segment).}
    \vspace{-5mm}
    \label{tab:ablation2}  
        \begin{tabular}{l c c c c c}
            \toprule
            & \textbf{Revenue} & \textbf{Impression} & \textbf{CTR} & \textbf{CTR30} & \textbf{CVR} \\
            \midrule
            \textbf{R2} & -1.43\% & -2.41\% & +15.87\% & +15.24\% & - \\
            \textbf{R3} & +0.21\% & +0.33\% & +0.97\% & -1.17\% & - \\
            \bottomrule
        \end{tabular}
    \vspace{-5mm}
\end{table}  

\textbf{R4}: In this configuration, we tailored the reward function specifically for conversion campaigns
by focusing exclusively on CVR and excluding other components in $Estimated\_User\_Value$. This adjustment aims at driving more
conversions by directly incentivizing the desired outcome.
As shown in Table~\ref{tab:ablation3}, this strategy resulted in
an increase of $+0.28\%$ of CVR, despite a slight
decrease of CTR gains for conversion campaigns compared with \textbf{R3}. These metrics remain strong, suggesting
that while click-based engagement is marginally reduced, the overall focus on conversion continues to be successful.

\begin{table}[h]  
    \centering  
    \caption{Comparing R3 and R4 on conversion campaigns (treated segment).}
    \vspace{-3mm}
    \label{tab:ablation3}  
        \begin{tabular}{l c c c c c}
            \toprule
            & \textbf{Revenue} & \textbf{Impression} & \textbf{CTR} & \textbf{CTR30} & \textbf{CVR} \\
            \midrule
            \textbf{R3} & +1.20\% & -2.16\% & +10.11\% & +6.05\% & -1.09\% \\
            \textbf{R4} & +0.17\% & -0.35\% & +8.28\% & +5.14\% & +0.28\% \\
            \bottomrule
        \end{tabular}
\end{table}  
  
To summarize, these ablation studies demonstrate that by meticulously crafting and fine-tuning reward definitions to match specific objectives, the metrics for each campaign type can be enhanced. This approach is particularly effective for multi-objective optimization problems, as it enables the system to balance various factors according to the established reward criteria.

\subsection{Predicted Hyperparameters}
In this section, we visualize the predicted hyperparameters to gain deeper insights into the model's behavior.

We categorize users into buckets according to their historical CTR and then calculate the average of the predicted hyperparameters for each bucket. Specifically, we concentrate on the predicted weights $w_{click}$, $w_{conversion}$, and the reserve price $b$. Similarly, users can also be grouped into buckets based on their historical CVR and revenue per impression (CPM).

The results are summarized in Figure~\ref{fig:visualization}. The top row illustrates the outcomes when categorizing by historical CTR. It shows that the model adjusts by increasing $w_{click}$ and decreasing $w_{conversion}$ and the reserve price $b$ for users with a higher likelihood of clicking. This suggests that the model effectively learned to recommend highly engaging content to these users, focusing less on conversion and revenue. This behavior strongly indicates that the proposed method successfully achieved personalization while optimizing for the predefined reward function. Similar conclusions can be drawn from the middle and bottom rows when categorizing by user historical CVR and CPM, respectively. An intriguing observation is that the predicted reserve price $b$ tends to be higher when the user has a higher historical CVR, compared to when the user has a higher historical CTR. Our hypothesis is that advertisers typically value conversions (e.g., purchases, sign-ups) more than clicks. A user with a higher historical CVR is more likely to convert after clicking on an ad, making them more valuable to advertisers. The system sets higher reserve prices for these users to maximize revenue from advertisers targeting conversions.

\section{Conclusion and Future Work}
In conclusion, this work introduces a novel deep reinforcement learning framework designed to address the complex challenge of multi-objective ranking utility tuning. Our approach stands out by providing an automatic, adaptive, and personalized solution, which is crucial for optimizing performance in dynamic environments. The effectiveness of our framework was validated through its application to Pinterest’s ads ranking system in a production setting, demonstrating substantial performance improvements, underscoring the practical viability and impact of our model in real-world ad recommender systems. Furthermore, the visualization of model predictions revealed that the model adeptly learns to leverage user features, thereby achieving personalized hyperparameter tuning. These findings highlight the potential of our framework to enhance the efficiency and effectiveness of ad ranking systems, paving the way for future research and development in this domain.

We also outline several avenues for future research and development based on the limitations and challenges identified in our current study. (1) On-Policy Reinforcement Learning and Stability: Our current approach involves behavior policy logging that is randomly sampled, categorizing the problem as an off-policy reinforcement learning (RL) problem. However, when deploying the proposed methods in a production environment, it is crucial for the model to learn from the data it generates itself, transitioning to an on-policy framework. Future work should focus on ensuring the stability of the model in such a setting. This includes developing robust mechanisms for continuous learning and adaptation, as well as implementing comprehensive monitoring systems to detect and address potential instabilities or drifts in model performance. (2) Incorporating Long-Term Rewards: The current reward definition in our model primarily considers immediate, short-term rewards. This approach may overlook the potential long-term effects and benefits of certain actions. Future research should explore methodologies for modeling long-term rewards, which could involve the integration of temporal discounting or the development of new reward structures that better capture the delayed benefits of actions.

\section{Acknowledgements}

We would like to thank Jay Adams for his insightful discussions on DRL-PUT.


\newpage

\bibliographystyle{ACM-Reference-Format}
\bibliography{l1utility}

\end{document}